\documentclass{article}

\usepackage[preprint,nonatbib]{neurips_2021}

\usepackage[utf8]{inputenc} 
\usepackage[T1]{fontenc}    
\usepackage{hyperref}       
\usepackage{url}            
\usepackage{booktabs}       
\usepackage{amsfonts}       
\usepackage{nicefrac}       
\usepackage{microtype}      
\usepackage{xcolor}         

\usepackage{amssymb}
\usepackage{amsmath,tabu,amsthm}
\usepackage{mathtools}
\usepackage{color}
\usepackage{lscape}

\usepackage{tikz}
\usepackage{tikz-cd}
\usetikzlibrary{decorations.markings}
\usetikzlibrary{decorations.pathreplacing}
\usepackage{enumitem}
\usepackage{mathtools}
\setenumerate[1]{label=\arabic*.}
\usepackage{bbm}
\usepackage{pifont}
\usepackage[numbers]{natbib}

\newcommand{\BaseSpace}{\ensuremath{I}}
\newcommand{\TargetSpace}{\ensuremath{Y}}
\newcommand{\Dataset}{\ensuremath{D}}
\newcommand{\Subbase}{\ensuremath{U}}
\newcommand{\Top}{\ensuremath{\mathcal{T}}}

\newcommand{\Assign}{\ensuremath{A}}
\newcommand{\DataSheaf}{\ensuremath{\mathcal{D}}}
\newcommand{\res}{\ensuremath{\text{res}}}
\newcommand{\Models}{\ensuremath{M}}
\newcommand{\ModelSheaf}{\ensuremath{\mathcal{M}}}
\newcommand{\ModelMap}{\ensuremath{\Phi^\ModelSheaf}}
\newcommand{\Incon}{\ensuremath{\text{Incon}}}
\newcommand{\xmark}{\ding{55}}
\newcommand{\Graff}{\text{Graff}}

\DeclareMathOperator*{\argmin}{arg\,min}

\newcommand{\omitt}[1]{}

\newtheorem{theorem}{Theorem}[section]

\newtheorem{proposition}[theorem]{Proposition}

\newtheorem{remark}[theorem]{Remark}

\theoremstyle{definition}
\newtheorem{definition}{Definition}[section]
\newtheorem{example}{Example}

\title{Sheaves as a Framework for Understanding and Interpreting Model Fit}

\author{%
  Henry~Kvinge \\
  Pacific Northwest National Laboratory\\
  \vspace{2mm}
  Seattle, WA, USA\\
  Department of Mathematics\\
  University of Washington\\
  Seattle, WA, USA\\
  \texttt{henry.kvinge@pnnl.gov} \\
  \And
  Brett~Jefferson \\
  Pacific Northwest National Laboratory \\
  Seattle, WA, USA\\
  \texttt{brett.jefferson@pnnl.gov} \\
  \AND
  Cliff Joslyn \\
  Pacific Northwest National Laboratory \\
  Seattle, WA, USA\\
  \texttt{cliff.joslyn@pnnl.gov} \\
  \And
  Emilie Purvine \\
  Pacific Northwest National Laboratory \\
  Seattle, WA, USA\\
  \texttt{emilie.purvine@pnnl.gov} \\
}

\begin{document}
\maketitle

\begin{abstract}
As data grows in size and complexity, finding frameworks which aid in interpretation and analysis has become critical. This is particularly true when data comes from complex systems where extensive structure is available, but must be drawn from peripheral sources. In this paper we argue that in such situations, sheaves can provide a natural framework to analyze how well a statistical model fits at the local level (that is, on subsets of related datapoints) vs the global level (on all the data). The sheaf-based approach that we propose is suitably general enough to be useful in a range of applications, from analyzing sensor networks to understanding the feature space of a deep learning model.
\end{abstract}

\section{Introduction}
\label{sect-intro}

Data is being collected at an ever-increasing rate in an ever-broader range of modalities. It is more and more frequently the case that extracting useful information from such large datasets requires the integration of sophisticated analytical techniques in combination with deep domain knowledge (often drawn from independent databases). While progress has been made in developing approaches to better visualize and explore the data itself, techniques for bringing outside knowledge into the analysis of the statistical models built on top of this data either remain mostly ad hoc or have not caught up to the size or scope of current state-of-the-art models. In this paper we propose a sheaf-theoretic approach to address this challenge.


Despite their ubiquity in mathematics, sheaves have only recently started playing a role in data science, where they have been leveraged as data structures which can systematically capture information from many non-independent data-streams. Sheaf frameworks have been developed for uncertainty quantification in geolocation \cite{joslyn2019sheaf}, air traffic control monitoring \cite{mansourbeigi2017sheaf}, and learning signals on graphs \cite{hansen2019learning}. Much of the inspiration for the present work comes from sheaf-theoretic constructions meant to analyze sensor networks \cite{robinson2017sheaves}, \cite{curry2013sheaves}. We use these ideas to analyze the fit of data-driven models. 

Our motivation arises from the observation that the quality of a model's ``fit'' may vary dramatically across subpopulations of a dataset. Indeed, it is increasingly apparent that the construction of models that are both robust and often highly-overparametrized requires understanding model performance not only at the global level (for example, the accuracy over an entire dataset) but also at the local level (precision on meaningful subsets of the data) \cite{sohoni2020no}. While our focus is not on overparametrized models in this work, we do give an example of how our framework might be applied to this setting in Section \ref{subsectlatentspace}.

To define sheaves on a dataset we first construct a topology based on metadata, with open sets corresponding to subsets of related points. We build a sheaf and presheaf on top of this topology. We call the sheaf the {\emph{data sheaf}} $\DataSheaf$. It consists of all possible data value observations. Then the {\emph{model presheaf}} $\ModelSheaf$ consists of a specified family of functions associated with each open set. For example, each data point might be indexed by a spatial location on Earth, an open set $U$ might then consist of spatial locations that are nearby each other, and a data {\emph{observation}} might be a measurement of temperature and wind speed. Then a section of the data sheaf at $U$ is a function $f:U \rightarrow \mathbb{R}^2$ which associates an element of $\mathbb{R}^2$ (temperature and wind speed) to each spatial location in $U$. We might choose $\ModelSheaf(U)$ to consist of all $1$-dimensional affine subspaces (lines) in $\mathbb{R}^2$. The process of modeling the points in $U$ as a line in $\mathbb{R}^2$ is equivalent to defining a map from the space of sections $\DataSheaf(U)$ to the space of sections $\ModelSheaf(U)$.


As suggested by this example, we identify a method of modeling data on each open set $U$ in our topology with a map $\ModelMap_U: \DataSheaf(U) \rightarrow \ModelSheaf(U)$. That is, for an observation of data on $U$ we have a rule for picking a model. In general, $\ModelMap$ will not be a presheaf morphism. Inspired by the notion of consistency from \cite{robinson2017sheaves}, we introduce a family of statistics, model map {\emph{inconsistency}}, which take values in $\mathbb{R}_{\geq 0}$ and measure the extent to which $\ModelMap$ is not a presheaf morphism. Indeed, we show in Proposition \ref{prop-inconsistency-presheaf} that $\ModelMap$ is a presheaf morphism if and only if the inconsistency of $\ModelMap$ is $0$.

Model map inconsistency is important because it allows us to point to specific subpopulations of a dataset on which a given model's ability to fit the data changes. This is critical because in the real world, the summary statistics used in academic benchmark studies often do not provide sufficient feedback on model behavior and performance. Indeed, we need to understand specific, systemic failures in a model before it is deployed. Further, note that our presheaf structure can also capture specific statistics associated with a global model. The simplest example of this might be the accuracy of a predictive model across various subpopulations of a test set. In this case ``model'' in the term model presheaf, might more appropriately be called ``model performance presheaf''. To illustrate this latter use, we end this work by using inconsistency statistics to probe the feature space of a ResNet50 convolutional neural network \cite{he2016deep} that has been trained on the large image database ImageNet \cite{russakovsky2015imagenet}. We show how our sheaf-theoretic framework can be used to illuminate biases in rich and complex computer vision models such as this. 


\section{Datasets and models as sheaves}

\subsection{Notation and underlying topology}
\label{sect-notation-topology}

Below we assume that set $\BaseSpace$ indexes the elements of the dataset $\Dataset$ that we will be working with so that
\begin{equation*}
    \Dataset = \{x_i \;|\: i \in I\},
\end{equation*} 
and each $x_i \in \Dataset$ takes a value in {\emph{target space}} $\TargetSpace$. We could equivalently encode $D$ as a function $f_\Dataset: \BaseSpace \rightarrow \TargetSpace$, where $f_\Dataset(i) = x_i$. This interpretation will be useful when defining data sheaves in Section \ref{sect-data-sheaf}. 




\begin{table}
{\def\arraystretch{1.5}%
\begin{center}
{\renewcommand{\arraystretch}{1.2}
    \begin{tabular}{cc}
        Researcher & Publications\\\hline
        $a$ & 5\\
        $b$ & 6\\
        $c$ & 8\\
        $d$ & 7\\
        $e$ & 4\\
        $f$ & 5\\\hline
    \end{tabular}}
\end{center}}
\caption{Number of publications per researcher in Example \ref{ex-set-up-framework}.\ref{ex-set-up-framework-ex-1}}
\label{table-researcher-pubs}
\end{table}

\begin{example} \label{ex-set-up-framework}  
\begin{enumerate}

\item \label{ex-set-up-framework-ex-1} (Toy example) Let $D$ be a dataset that consists of the number of publications for $6$ researchers. Denote these researchers by $a,b,c,d,e,f$. The number of publications of each is recorded in Table \ref{table-researcher-pubs}. Then $\BaseSpace = \{a,b,c,d,e,f\}$ and $\TargetSpace = \mathbb{R}$ (we use $\mathbb{R}$ rather than $\mathbb{N}$ because we will later want to take averages in the target space).



\item (Sensors) Suppose that $\Dataset$ consists of real-valued readings from a collection of distinct sensors. Then $\BaseSpace$ indexes the set of all sensors, $\TargetSpace = \mathbb{R}$, and $f_\Dataset: I \rightarrow \mathbb{R}$ is a function that to each sensor assigns its reading.

\item (Gene expression) If $\Dataset$ is a gene expression level dataset where readings are taken from each gene at a fixed set of  $r$ time steps, then $\BaseSpace$ is a set that indexes all genes whose expression level are being measured, $\TargetSpace = \mathbb{R}^r$, and $f_\Dataset: I \rightarrow \mathbb{R}^r$ assigns to each gene a vector recording its $r$ readings.

\item (Computer vision feature extractor) Let $J \subset \mathbb{R}^{h \times w \times 3}$ be a collection of height $h$ and width $w$ RGB images and let $\varphi$ be a convolutional neural network (CNN) feature extractor that has been trained to map images from their usual pixel space, to vectors in a feature space $\mathbb{R}^r$ where spatial relationships can be related to image content (that is, if $x_1,x_2 \in \mathbb{R}^{h \times w \times 3}$ are two images and $||\varphi(x_1) - \varphi(x_2)||_{\ell_2}$ is small, then there is a high likelihood that $x_1$ and $x_2$ contain similar content). Feature extractors such as $\varphi$ are an important component in many state-of-the-art methods in computer vision (see for example \cite{snell2017prototypical}). 


To better understand the features extracted by $\varphi$, one might be motivated to analyze $\Dataset = \varphi(J)$. In this setting $\BaseSpace$ is the list of all images in $J$, $\TargetSpace = \mathbb{R}^r$, and $f_\Dataset$ maps $i \in \BaseSpace$ to $\varphi(x_i)$. We analyze a specific example of this set-up in Section \ref{subsectlatentspace}.
\end{enumerate}
\end{example}

In order to put a topology on $\BaseSpace$, $\BaseSpace$ must have some additional structure. Thus we assume that there exist subsets $\Subbase_1, \Subbase_2, \dots, \Subbase_k$ of $\BaseSpace$ that capture relationships between elements of $\BaseSpace$. This is the external information described in Section \ref{sect-intro}. We do not assume that $\Subbase_1, \dots, \Subbase_k$ are disjoint. 

\begin{example}
We give some possible $U_1, \dots, U_k$ for each part of Example \ref{ex-set-up-framework}.
\begin{enumerate}
    \item (Toy example) Let $\Subbase_1 = \{a,b,c,d\}$ and $\Subbase_2 = \{c,d,e,f\}$ be known collaborations between researchers $a,b,c,d,e,f$.
    \item (Sensors) For each $1 \leq p \leq k$ let $\Subbase_p$ be the subset of $\BaseSpace$ that contains all sensors of a certain type or measuring a given modality. Alternatively, let each $\Subbase_p$ contain indices corresponding to sensors located in a given region.
    \item (Gene expression) Subset $\Subbase_p$ might contain all genes encoding proteins involved in a specific biochemical pathway $p$. 
    \item (Features extracted by a CNN) If the images in $J$ are labeled based on whether they contain any of $r$ different classes of object, then $\Subbase_p$ might be the subset of $\BaseSpace$ indexing all images with a specific object in it. For example, $x_i \in J$ might be an image containing a cat and a ball of yarn, in which case $i \in U_{\text{cat}}$ and $i \in U_{\text{yarn}}$.
\end{enumerate}
\end{example}

Our proposed sheaf-theoretic framework requires three components: (i) a {\emph{topology}} $\Top_B$ on $\BaseSpace$ built using $B = \{\Subbase_1, \Subbase_2, \dots, \Subbase_k\}$, (ii) a {\emph{data sheaf}} $\DataSheaf$ on $\BaseSpace$ where the raw data from datasets $\Dataset$ is stored as a choice of sections, and (iii) a {\emph{model presheaf}} $\ModelSheaf$ also on $\BaseSpace$ which defines the type of model whose fit we want to understand.

Let $\Top_B$ be the topology on $\BaseSpace$ generated by subbasis $B = \{U_1, \dots, U_k\}$. Note that $\Top_B$ captures a notion of space on the elements of dataset $D$ based on the external information found in $B$. In this paper $\Top_B$ will underlie all the sheaves we construct unless otherwise specified. Since $\BaseSpace$ is finite, then $\Top_B$ is a finite topology. 

In Section \ref{sect-local-consistency} we will compare sections of sheaves between open sets. In order to do this systematically we will take advantage of a lattice-theoretic viewpoint of $\Top_B$. The open sets defining $\Top_B$ have a poset structure, $(\Top_B,\leq)$, based on set inclusion. That is, for $U,V \in \Top_B$, $V \leq U \Leftrightarrow V \subseteq U$.
In fact, $\Top_B$ is a lattice since any two open sets $U$ and $V$ always have a meet and join given by their intersection and union respectively. For fixed open set $U$ in $\Top_B$, consider the order ideal generated from $U$, $\Lambda_U := \{V \in \Top_B\;|\;V \subseteq U \}$.
Since $\Top_B$ is finite, and hence $\Lambda_U$ is finite, there exists a finite filtration on $\Lambda_U$, 
\begin{equation*}
    \{U\} = \Lambda_U^0 \subset \Lambda_U^1 \subset \dots \subset \Lambda_U^k = \Lambda_U
\end{equation*}
where $\Lambda_U^i$ consists of all those $V \in \Lambda_U$ such that there is a maximal chain in $\Lambda_U$, $V = V_j \subset V_{j-1} \subset \dots  \subset V_1 \subset U$ with $j \leq i$. 

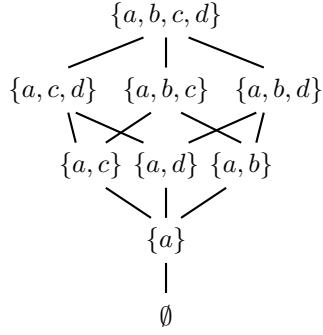
\begin{figure}[pt!]
\begin{center}
\begin{tikzpicture}
\node at (0,4) {$\{a,b,c,d\}$};
\node at (1.5,3) {$\{a,b,d\}$};
\node at (0,3) {$\{a,b,c\}$};
\node at (-1.5,3) {$\{a,c,d\}$};
\node at (-1,2) {$\{a,c\}$};
\node at (0,2) {$\{a,d\}$};
\node at (1,2) {$\{a,b\}$};
\node at (0,1) {$\{a\}$};
\node at (0,0) {$\emptyset$};

\draw[thick] (1.3,3.3)--(.3,3.7);
\draw[thick] (-1.3,3.3)--(-.3,3.7);
\draw[thick] (0,3.3)--(0,3.7);
\draw[thick] (.2,2.7)--(1,2.3);
\draw[thick] (1.2,2.7)--(.3,2.3);
\draw[thick] (-1.2,2.7)--(-.3,2.3);
\draw[thick] (1.2,2.3)--(1.3,2.7);
\draw[thick] (-.8,2.3)--(-.2,2.7);
\draw[thick] (-1.2,2.3)--(-1.3,2.7);
\draw[thick] (.2,1.3)--(.8,1.7);
\draw[thick] (0,1.3)--(0,1.7);
\draw[thick] (-.2,1.3)--(-.8,1.7);
\draw[thick] (0,0.3)--(0,.7);
\end{tikzpicture}
\end{center}
\caption{The Haase diagram of the finite topology from Example \ref{ex-lattice-filtration}.}
\label{fig-Haase-diagram}
\end{figure}

\begin{example} \label{ex-lattice-filtration}
Consider the topology, depicted as a Haase diagram in Figure \ref{fig-Haase-diagram}. Let $U = \{a,b,d\}$. Then 
\begin{equation*}
    \Lambda_U = \big\{U, \{a,d\},\{a,b\},\{a\},\emptyset\}
\end{equation*}
with 
\begin{align*}
&\Lambda_U^0 = \{U\},\\ 
&\Lambda_U^1 = \big\{U,\{a,d\},\{a,b\}\big\},\\
&\Lambda_U^2 = \big\{U,\{a,d\},\{a,b\},\{a\}\big\},\\
&\Lambda_U^3 = \big\{U,\{a,d\},\{a,b\},\{a\},\emptyset\big\}.
\end{align*}
\end{example}


\subsection{Data sheaves and model presheaves}
\label{sect-data-sheaf}

Our goal in this section is to realize our dataset $\Dataset$ as the section of a sheaf on the topological space $\Top_B$ that we constructed in Section \ref{sect-notation-topology}. We point the reader to \cite[Part 1, Chapter 2]{vakil2017rising} for an introduction to presheaves and sheaves.

\begin{definition}
Let $\BaseSpace$ be a finite set with topology generated by subbasis $B = \{\Subbase_1, \dots, \Subbase_k\}$ and let $\TargetSpace$ be a set. Then the {\emph{data sheaf}} $\DataSheaf$ associated to $(\BaseSpace,B,\TargetSpace)$ is the sheaf that:
\begin{itemize}
    \item to each open set $U \subseteq \BaseSpace$ assigns the set $\DataSheaf(U)$ of all functions from $U$ to $\TargetSpace$,
    \item and to each open set $V \subseteq U$ assigns the usual restriction of functions map $\res^{\DataSheaf}_{U,V}: \DataSheaf(U) \rightarrow \DataSheaf(V)$ which restricts functions from $U \rightarrow Y$ to functions from $V \rightarrow Y$.
\end{itemize}
\end{definition}

\begin{remark}
Note that the fact that $\DataSheaf$ is a sheaf, rather than just being a presheaf, follows from elementary arguments that spaces of functions (without further constraints) satisfy the gluing and locality axiom.
\end{remark}


\begin{figure}
\begin{center}
\includegraphics[width=.5\columnwidth]{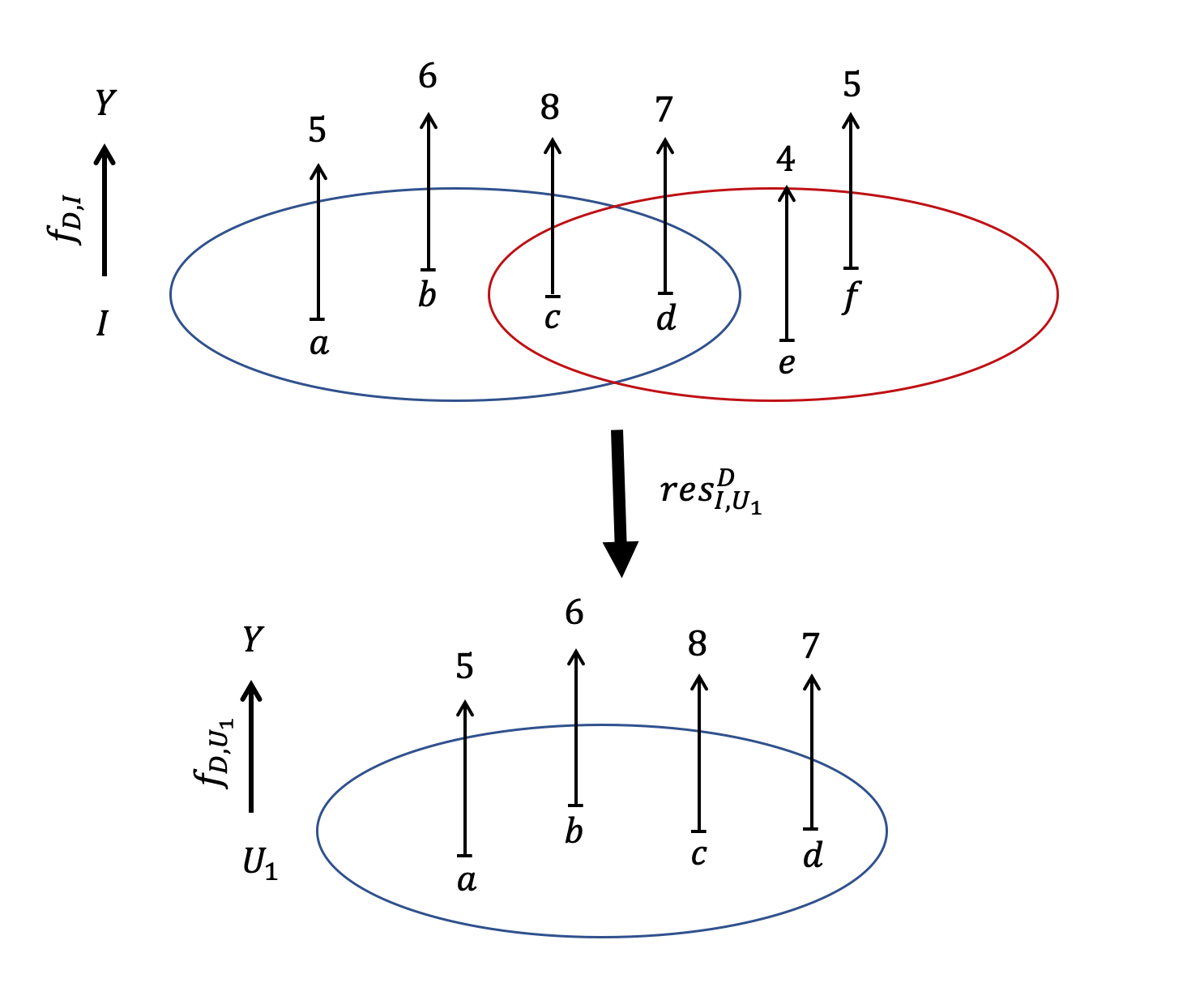}
\end{center}
\caption{A visualization of the restriction map $\res_{\BaseSpace,U_{1}}^{\DataSheaf}$ that takes a section $f_\Dataset$ from $\DataSheaf(\BaseSpace)$ to $f_{\Dataset,U_1}$ in $\DataSheaf(U_1)$.}
\label{fig-toy-diagram}
\end{figure}

\begin{remark}
To make our constructions applicable to a large range of data science settings, we choose to work with sheaves taking values in the category $\mathbf{Set}$. For specific problems, one can often choose to work with sheaves taking values in a category with more structure.
\end{remark}

Recall that we can realize $\Dataset$ as a function $f_\Dataset: \BaseSpace \rightarrow Y$ which takes an element $i \in \BaseSpace$ as input and returns the measured value $f_\Dataset(i)$ at that point. The same construction holds for any subsets of $I$ in the obvious way. Thus for any open set $U$ of $I$, there is a map that takes a dataset $D$ with index set $\BaseSpace$ to a section of $\DataSheaf(U)$, $f_{\Dataset,U}:U \rightarrow Y$. Together these define the map,
\begin{equation}
\label{eqn-data-to-assignment}
    \Dataset \mapsto \{f_{\Dataset,U}: U \rightarrow Y \;|\; U \in \Top_B\}.
\end{equation}
From this perspective $f_\Dataset$ is a global section of $\DataSheaf$, in particular, $f_{\Dataset,\BaseSpace}$ in $\DataSheaf(\BaseSpace)$.


Following \cite{robinson2017sheaves}, if $\mathcal{S}$ is a sheaf (or presheaf) on space $X$ with topology $\Top$, we call a choice of section $a_U$ from each open set $U \in \Top$,
\begin{equation*}
    \Assign = \prod_{U \in \Top}a_U,
\end{equation*}
an {\emph{assignment}} on $\mathcal{S}$. We note that any global section $a_X$ of $\mathcal{S}$ induces an assignment by setting $a_U = \res_{X,U}(a_X)$ for each $U \in \Top$. 

$\Dataset$ defines an assignment on $\DataSheaf$ via global section $f_D$, $F_\Dataset = (f_{\Dataset,U})_{U \in \Top_B}$.

\begin{example} \label{ex-publication-data-sheaf}
(Toy example) There are $5$ open subsets in the topology generated by subbasis $\Subbase_1 = \{a,b,c,d\}$ and $\Subbase_2 = \{c,d,e,f\}$ on $\BaseSpace = \{a,b,c,d,e\}$: $\emptyset, \{c,d\}, \Subbase_1, \Subbase_2, \BaseSpace$.
The dataset $D$ listed in Table \ref{table-researcher-pubs}, defines a function on each of these subsets. For example, $f_{\Dataset,\Subbase_1}: \Subbase_1 \rightarrow \mathbb{R}$, an element of $\DataSheaf(\Subbase_1)$, is a function that sends $f(a) = 5, \;\; f(b) = 6, \;\; f(c) = 8, \;\; f(d) = 7$. On the other hand, $f_{\Dataset,\{c,d\}}: \{c,d\} \rightarrow \mathbb{R}$ is the function that sends, $f(c) = 8$ and $f(d) = 7$.
The full assignment $F_\Dataset$ on $\DataSheaf$ induced by $D$ consists of $f_{\Dataset,\BaseSpace}, f_{\Dataset,\Subbase_1},f_{\Dataset,\Subbase_2},f_{\Dataset,\{c,d\}},f_{\Dataset,\emptyset}$. Figure \ref{fig-toy-diagram} contains a visualization of the restriction map $\res^{\DataSheaf}_{\BaseSpace,U_1}$ taking $f_{\Dataset,\BaseSpace}$ to $f_{\Dataset,U_1}$.
\end{example}
 
Note that $F_\Dataset$ contains a significant amount of redundant information, being completely determined by its single global section $f_\Dataset = f_{\Dataset,\BaseSpace}$. On the other hand, one can easily find assignments $\Assign = (a_U)_{U \in \Top_B}$ that are not determined by their associated global section $a_\BaseSpace$.

\begin{table}
{\def\arraystretch{1.5}%
\begin{center}
    \begin{tabular}{ccccccc}
        Section & $a$ & $b$ & $c$ & $d$ & $e$ & $f$\\\hline
        $f_\BaseSpace$ & 4 & 4 & 2 & 3 & 2 & 5\\
        $f_{U_1}$ & 4 & 4 & 2 & 3 & \xmark & \xmark\\
        $f_{U_2}$ & \xmark & \xmark & 2 & 3 & 2 & 5\\
        $f_{\{c,d\}}$ & \xmark & \xmark & 2 & 3 & \xmark & \xmark\\
        \hline
    \end{tabular} \quad\quad\quad
        \begin{tabular}{ccccccc}
        Section & $a$ & $b$ & $c$ & $d$ & $e$ & $f$\\\hline
        $f_\BaseSpace$ & 4 & 4 & 2 & 3 & 2 & 5\\
        $f_{U_1}$ & 3 & 2 & 2 & 4 & \xmark & \xmark\\
        $f_{U_2}$ & \xmark & \xmark & 5 & 6 & 0 & 2\\
        $f_{\{c,d\}}$ & \xmark & \xmark & 3 & 2 & \xmark & \xmark\\
        \hline
    \end{tabular}
\end{center}}
\caption{Examples of assignments on the data sheaf from the toy publication example that do correspond to a global assignment (left) and do not correspond to a global assignment (right).}
\label{ex-consistent-assignment}
\end{table}

\begin{example} \label{ex-consistent-inconsistent}
(Toy Example) We return to our publication example from Example \ref{ex-publication-data-sheaf}. The reader can check that while the assignment described on the left in Table \ref{ex-consistent-assignment} is induced by a single global section $f_{\Dataset,\BaseSpace}: \BaseSpace \rightarrow \mathbb{R}$ which assigns
\begin{align*}
    &f_\BaseSpace(a) = 4, \; f_\BaseSpace(b) = 4, \; f_\BaseSpace(c) = 2, \; f_\BaseSpace(d) = 3, \\
    &f_\BaseSpace(e) = 2, \; f_\BaseSpace(f) = 5.
\end{align*}
The assignment provided on the right in Table \ref{ex-consistent-assignment} is not induced by a global section. 
\end{example}





We give a special definition for those assignments that arise from a global section on a data sheaf (and hence from a dataset $\Dataset$).

\begin{definition}
Let $\mathcal{S}$ be a sheaf (or presheaf) on space $X$ with topology $\Top$. An assignment $\Assign = (a_U)_{U \in \Top}$ is said to be {\emph{consistent}} if for all $U,V \in \Top$ with $V \subseteq U$, $a_V = \res_{U,V}(a_U)$. Otherwise, $\Assign$ is said to be {\emph{inconsistent}}.
\end{definition}

The assignment defined on the left in Table \ref{ex-consistent-assignment} is consistent. The one on the right is inconsistent. It is clear that data assignment $F_D$ induced by a dataset $D$ is always consistent by construction. 

\subsection{The model presheaf and modeling map}
\label{sect-model-presheaf}

In this section we define a presheaf on $\BaseSpace$. While the data sheaf $\DataSheaf$ encodes instances of raw data collected on $\BaseSpace$, this new presheaf will encode local models of $\Dataset$. As in previous sections we assume that $\BaseSpace$ is the space that indexes data from some dataset $\Dataset$ and $\BaseSpace$ has topology $\Top_B$ generated by subbasis $B = \{\Subbase_1,\dots,\Subbase_k\}$.

\begin{definition}
\label{def-model-sheaf}
Let $\DataSheaf$ be a data sheaf on $\BaseSpace$. A {\emph{model presheaf}} on $\BaseSpace$ is a presheaf $\ModelSheaf$ on $\BaseSpace$ along with a {\emph{modeling map}} $\ModelMap = (\ModelMap_U)_{U \in \Top_B}$ which consists of functions $\ModelMap_U: \DataSheaf(U) \rightarrow \ModelSheaf(U)$ for each $U \in \Top_B$. For section $f_U \in \DataSheaf(U)$, we call $\ModelMap_U(f_U)$ a {\emph{model}} of $f_U$.
\end{definition}

\begin{remark}
Note that the modeling map $\ModelMap$ is a critical component of the model presheaf $\ModelSheaf$. In fact without the modeling map, $\ModelSheaf$ is just a presheaf with no additional structure. Further, $\ModelMap$ is generally not a presheaf morphism. In fact, we will derive a measure of global ``model inconsistency'' from the extent to which $\ModelMap$ fails to be a presheaf morphism.
\end{remark}

\begin{remark}
If we are handed a dataset, the data sheaf is generally implicit based on the form that the data takes. The space of sections associated with the model presheaf is chosen based on the type of models that one wants to use on the data. The least obvious choice in the framework that we present in this paper then is the definition of restriction maps in the model presheaf. In data science we rarely think about how to modify an existing model built to fit one dataset so that it instead fits a subset. When in doubt, we advocate keeping things simple. For open sets $V \subseteq U$, identity maps often work well provided that $\ModelSheaf(V) = \ModelSheaf(U)$.
\end{remark}

\begin{example}
Below we give two examples of model presheaves.
\begin{itemize}
\item \label{ex-average-sheaf} For each open set $U$ in $\BaseSpace$, let $\DataSheaf$ assign to $U$ the space of functions from $U$ to $\mathbb{R}$ (in other words, any dataset $\Dataset$ defined by a global section of this data sheaf consists of real-valued measurements indexed by elements of $\BaseSpace$). Then we define the {\emph{averaging presheaf}} $\ModelSheaf_{\text{avg}}$ to be the presheaf on $\BaseSpace$ such that 
\begin{equation*}
\ModelSheaf_{\text{avg}}(U) = \begin{cases}
\mathbf{0} & \text{if $U = \emptyset$}\\
\mathbb{R} & \text{otherwise}.
\end{cases}
\end{equation*}
with restriction maps given by
\begin{equation*}
    \res^{\ModelSheaf_{\text{avg}}}_{U,V} = \begin{cases}
    0 & \text{if $V = \emptyset$}\\
    id & \text{otherwise}\\
\end{cases}
\end{equation*}
for $U,V \in \Top_B$ with $V \subseteq U$. The model map $\Phi^{\ModelSheaf_{\text{avg}}}$ is defined such that for open set $U$ in $\BaseSpace$,
\begin{equation*}
    \Phi^{\ModelSheaf_{\text{avg}}}_U(f_{\Dataset,U})= \begin{cases}
    \frac{1}{|U|}\sum_{x \in U} f_{\Dataset,U}(x) & \text{if $U \neq \emptyset$}\\
    0 & \text{otherwise.}\\
\end{cases}
\end{equation*}
In other words, the average presheaf maps sections to their average. We could have chosen any number of different scalar valued statistics (for example, median, mode, maximum, etc.). 

\item For each open set $U$ in $\BaseSpace$, let $\DataSheaf$ assign to $U$ the space of functions from $U$ to $\mathbb{R}^r$. For integer $q < r$, let $\ModelSheaf_{\Graff(q,r)}$ be the constant presheaf such that
\begin{equation*}
\ModelSheaf_{\Graff(q,r)}(U) = \begin{cases}
\mathbf{0} & \text{if $U = \emptyset$}\\
\Graff(q,r) & \text{otherwise}
\end{cases}
\end{equation*}
where $\Graff(q,r)$ is the Grassmannian of $q$-dimensional affine subspaces in $\mathbb{R}^r$ \cite{graff} (note that this is not to be confused with the affine Grassmannian that appears more commonly in algebraic geometry and representation theory). For simplicity we assume that restriction maps are given by:
\begin{equation*}
    \res^{\ModelSheaf_{\Graff(q,r)}}_{U,V} = \begin{cases}
    0 & \text{if $V = \emptyset$}\\
    id & \text{otherwise.}\\
\end{cases}
\end{equation*}


There are numerous approaches to finding a $q$-dimensional affine subspace $W$ that best approximates vectors $v_1,\dots,v_m \in \mathbb{R}^r$ (assuming that $q < m$). Suppose that we have fixed such a method $G: \sqcup_{n \geq q} \underbrace{\mathbb{R}^r \times \dots \times \mathbb{R}^r}_{\text{$n$ times}} \rightarrow \Graff(q,r)$. Then we can define our model map $\Phi^{\ModelSheaf_{\Graff(q,r)}}$ such that for $U = \{i_1, \dots, i_\ell\}$ and $f_{\Dataset,U}$ we set 
\begin{equation*}
    \Phi^{\ModelSheaf_{\Graff(q,r)}}_U(f_{\Dataset,U})= \begin{cases}
    G(f_{\Dataset,U}(i_1),\dots, f_{\Dataset,U}(i_\ell)) & \text{if $U \neq \emptyset$}\\
    0 & \text{otherwise.}\\
\end{cases}
\end{equation*}
Note that for this construction to hold we need that if $U \neq \emptyset$ then $|U| \geq q$. In other words, $q$ should be chosen based on the size of the smallest open set. One can also choose to vary $q$ for different $U \in \Top_B$, but in this case the identity can no longer be used for certain restriction maps.

\end{itemize}
\end{example}

\begin{example}
\label{ex-first-models}
(Toy example) Suppose that we construct an averaging presheaf based on researcher publication numbers from Example \ref{ex-set-up-framework}.\ref{ex-set-up-framework-ex-1} Then to each open set (except $\emptyset$) in $\BaseSpace = \{a,b,c,d,e,f\}$ we associate $\mathbb{R}$. The model map $\ModelMap$ then sends: $f_{\BaseSpace} \mapsto 5.8\overline{3}$, $f_{\BaseSpace,\Subbase_1} \mapsto 6.5$, $f_{\BaseSpace,\Subbase_2} \mapsto 6$, $f_{\BaseSpace,\{c,d\}} \mapsto 7.5$. We see that the two researchers $c$ and $d$ who are involved in multiple collaborations have more publications (on average) than those that do not. The difference between the average number of publications for $c$ and $d$, $\ModelMap(f_{\BaseSpace,\{c,d\}})$, and the average number of publications for all researchers in $\BaseSpace$ is equal to
\begin{equation*}
|(\res_{\BaseSpace,\{c,d\}} \circ \ModelMap)(f_{\BaseSpace}) - (\ModelMap \circ \res_{\BaseSpace,\{c,d\}})(f_{\BaseSpace})|.
\end{equation*}
Note that this can also be interpreted as the extent to which $\ModelMap$ fails to be a presheaf morphism with respect to restriction from the whole space $\BaseSpace$ to open set $\{c,d\}$. We will explore this in more detail in Section \ref{sect-local-consistency}.
\end{example}

\section{Inconsistency of models}
\label{sect-local-consistency}

In Example \ref{ex-first-models} we saw that the inconsistency of the average number of publications across different subsets of collaborators could be identified with the extent to which restriction maps commute with the model map. This motivates the idea of model inconsistency. This notion was inspired by the idea of a consistency radius \cite[Definition 20]{robinson2017sheaves} in data fusion.

\begin{definition}
\label{def-inconsistency}
Let 
\begin{itemize}
\setlength\itemsep{0em}
\item $\Dataset$ be a dataset with elements indexed by $\BaseSpace$ and taking values in $\TargetSpace$,
\item $(\BaseSpace,\Top_B)$ be the topological space associated with $\BaseSpace$ and subbasis $B = \{U_1,\dots,U_k\}$, 
\item $\DataSheaf$ be the data sheaf on $\BaseSpace$, $\ModelSheaf$ be a model presheaf on $\BaseSpace$, and $\ModelMap$ be the model map taking spaces of sections from $\DataSheaf$ to spaces of sections in $\ModelSheaf$,
\item $F_\Dataset$ be the assignment associated with $\Dataset$,
\item and $d_U: \ModelSheaf(U) \times \ModelSheaf(U) \rightarrow \mathbb{R}_{\geq 0}$ be a metric for each $U \in \Top_B$.
\end{itemize}
Recall that for open set $U \subseteq \BaseSpace$, $\Lambda_U$ is the order ideal defined by $U$ (i.e. all open sets contained in $U$). The {\emph{local inconsistency of assignment $F_\Dataset$ with respect to $(\DataSheaf,\ModelSheaf)$ at $U$}} is defined to be
\begin{equation}
\label{eqn-local-inconsis}
    \Incon(F_\Dataset,U) := \max_{V \in \Lambda_U}d_V(\res^\ModelSheaf_{U,V}(\ModelMap_U(f_{\Dataset,U})),\ModelMap_V(f_{\Dataset,V}))
\end{equation}
if $\Lambda_U$ is non-empty and $\Incon(F_\Dataset,U) = 0$ otherwise. The {\emph{global inconsistency of $F_\Dataset$ with respect to $(\ModelSheaf,\DataSheaf)$}} is defined to be
\begin{equation*}
    \Incon(F_\Dataset) := \max_{U \in \Top_B} \Incon(F_\Dataset,U).
\end{equation*}
\end{definition}

Note that local inconsistency of assignment $F_\Dataset$ with respect to the model presheaf/model map pair $(\ModelSheaf,\ModelMap)$ is closely tied to the extent to which the diagram 
\begin{equation}
\label{tikz-commutative-diagram}
    \begin{tikzpicture}
    \node at (0,1.5) {$\DataSheaf(U)$};
    \node at (0,0) {$\DataSheaf(V)$};
    \node at (3,1.5) {$\ModelSheaf(U)$};
    \node at (3,0) {$\ModelSheaf(V)$};
    
    \draw[->] (0.5,0) -- (2.5,0);
    \draw[->] (0.5,1.5) -- (2.5,1.5);
    \draw[->] (0,1) -- (0,.5);
    \draw[->] (3,1) -- (3,.5);
    
    \node at (1.5,1.8) {$\ModelMap_U$};
    \node at (1.5,.3) {$\ModelMap_V$};
    
    \node at (-.5,.75) {$\res^\DataSheaf_{U,V}$};
    \node at (3.55,.75) {$\res^\ModelSheaf_{U,V}$};
    \end{tikzpicture}
\end{equation}
fails to commute. We formalize this in a proposition. 

\begin{proposition}
\label{prop-inconsistency-presheaf}
As above, let $(X,\Top_B)$ be a topological space, $\DataSheaf$ a data sheaf, $\ModelSheaf$ a model presheaf  with model map $\ModelMap: \DataSheaf \rightarrow \ModelSheaf$, and $(d_U)_{U \in \Top}$ be a collection of metrics. Then $\ModelMap$ is a presheaf morphism if and only if for any consistent assignment $\Assign = (f_U)_{U \in \Top_B}$ of $\DataSheaf$, the local inconsistency of $A$ at any open set $U$ is always $0$. 
\end{proposition}

One direction of this proposition follows from the definition of a presheaf morphism (specifically the commutativity of restriction maps with each map $\ModelMap_U$ between the spaces of sections). The other direction uses the fact that a section of a data sheaf can always be extended to a global section $F$ on $\DataSheaf(\BaseSpace)$. A full proof can be found in the Appendix.

\omitt{
\begin{proposition}
Let  $(\BaseSpace,\Top_B), \DataSheaf, \ModelSheaf, \ModelMap$, and $d_V$ be as in Definition \ref{def-inconsistency}. $\ModelMap$ is a presheaf morphism if and only if for all assignments $\Assign = (a_U)_{U \in \Top_B}$ of $\DataSheaf$, $\Assign$ has global inconsistency $0$ with respect to $\ModelMap$. 
\end{proposition}
}


\begin{remark} 
Note that we have chosen to define the local consistency of a model presheaf $\ModelSheaf$ with respect to section $f_U$ by comparing $\res^{\ModelSheaf}_{U,V}\ModelMap_U(f_U)$ (the restriction of $\ModelMap_U(f_U)$) against $\ModelMap_V(f_V)$ for each $V$ such that $V \subset U$. We could have conversely chosen to compare $\ModelMap_U(f_U)$ against $\res^{\ModelSheaf}_{W,U}\ModelMap_W(f_W)$ for each $W$ such that $U \subset W$. In the former case, if there are no proper, non-empty open sets $V$ in $U$, then by definition the local inconsistency at $U$ is $0$ for all assignments. In the latter case the situation would be reversed and global sections would always have $0$ inconsistency. We felt the idea of a model having lower inconsistency when fitted on smaller subsets aligns better with notions of fit from machine learning.
\end{remark}


\begin{example}
\label{ex-pub-inconsistency}
(Toy example) We can compute the inconsistency of the consistent assignment in Table \ref{table-researcher-pubs} with respect to the average model presheaf. We only need to define a distance function $d_U$ on each copy of $\mathbb{R}$. We do this by choosing $d_U$ to be the standard distance between real numbers: $d_U(x,y) = |x-y|$, for $x,y \in \mathbb{R}$. The local inconsistencies (omitting the empty set) are as follows: $\Incon(A,\BaseSpace) = 3.67$, $\Incon(A,\Subbase_1) = 1$, $\Incon(A,\Subbase_2) = 1.5$, $\Incon(A,\{c,d\})= 0$. The global inconsistency is $\Incon(A) = 3.67$. This occurs at the section on the total space $\BaseSpace$. This makes sense since we would expect a single statistic to represent a smaller set of numbers better than a larger set of numbers.
\end{example}


\omitt{
\begin{proof}
Both parts of this proposition follow directly from the definition of  a presheaf morphism. For part  \ref{prop-1a}, we need only recall that if $\ModelMap$ is a presheaf morphism then the diagram in \eqref{tikz-commutative-diagram} commutes and hence $\res^\ModelSheaf_{U,V}(\ModelMap_U(f_U)) = \ModelMap_V(f_V)$ for any $V \subseteq U$ and any possible $f_U$.
\end{proof}
}

Assume that $\Top_B$ is a finite topology. Recall from Section \ref{sect-notation-topology} that for any $U \in \Top_B$ we can form the order ideal $\Lambda_U$, which is filtered such that $\{U\} = \Lambda_U^0 \subset \Lambda_U^1 \subset \dots \subset \Lambda_U^k = \Lambda_U$ for sufficiently large integer $k$. As illustrated in Example \ref{ex-pub-inconsistency}, it is often the case that the maximum value of
\begin{equation*}
d_V(\res^\ModelSheaf_{U,V}(\ModelMap_U(f_{\Dataset,U})),\ModelMap_V(f_{\Dataset,V}))    
\end{equation*}
 from \eqref{eqn-local-inconsis} is achieved for the smallest $V$ contained in $U$, since in this case $V$ and $U$ have maximum difference (that is, $U\setminus V$ is maximally large). Informally, inconsistency is often maximized by restricting to the ``smallest'' open set $V$ contained in $U$. In order to be able to ``see past'' this phenomenon, we introduce a final version of inconsistency which allows us to compare lack of commutativity of \eqref{tikz-commutative-diagram} only on ``nearby'' sets in the lattice associated with $\Top_B$.


\begin{definition}
Given the assumptions in Definition \ref{def-inconsistency}, if $U$ is an open set in $\Top_B$ and $\Lambda_U^1, \Lambda_U^2, \dots, \Lambda_U^k$ is the $\mathbb{Z}$-filteration on the order ideal $\Lambda_U$, then the {\emph{$j$-filtered inconsistency}} is defined as
\begin{equation}
\label{eqn-local-inconsis-filtered}
    \Incon_j(A,U) := \max_{V \in \Lambda^j_U}d_V(\res^\ModelSheaf_{U,V}(\ModelMap_U(f_U)),\ModelMap_V(f_V)).
\end{equation}
\end{definition}

\omitt{
\section{Computations in practice}

When the set $B = \{\Subbase_1,\dots,\Subbase_k\}$ is even moderately large, the topology $\Top_B$ generated by $B$ can be huge. In many cases, the contextual meaning held by sets in $B$ will be lost for many of the open sets in $\Top_B$. For example if, as in our genes/pathways example from Example \ref{}, $\BaseSpace$ is a set of genes and $B = \{\Subbase_1, \dots, \Subbase_{100}\}$ are subsets of genes whose proteins code for a particular pathway, then $\Top_B$ may not only be very large, but it may be hard to attach meaning to an element such as 
\begin{equation*}
    (\Subbase_1 \cup \Subbase_{10} \cup \Subbase_{23}) \cap (\Subbase_8 \cup \Subbase_{78}) \cap (\Subbase_{56} \cup \Subbase_{91}).
\end{equation*}
In this section we there define the notion of 
}

\subsection{Example: Analyzing the feature space of a convolutional neural network}
\label{subsectlatentspace}

\begin{figure*}[t]
\centering
\includegraphics[width=\columnwidth]{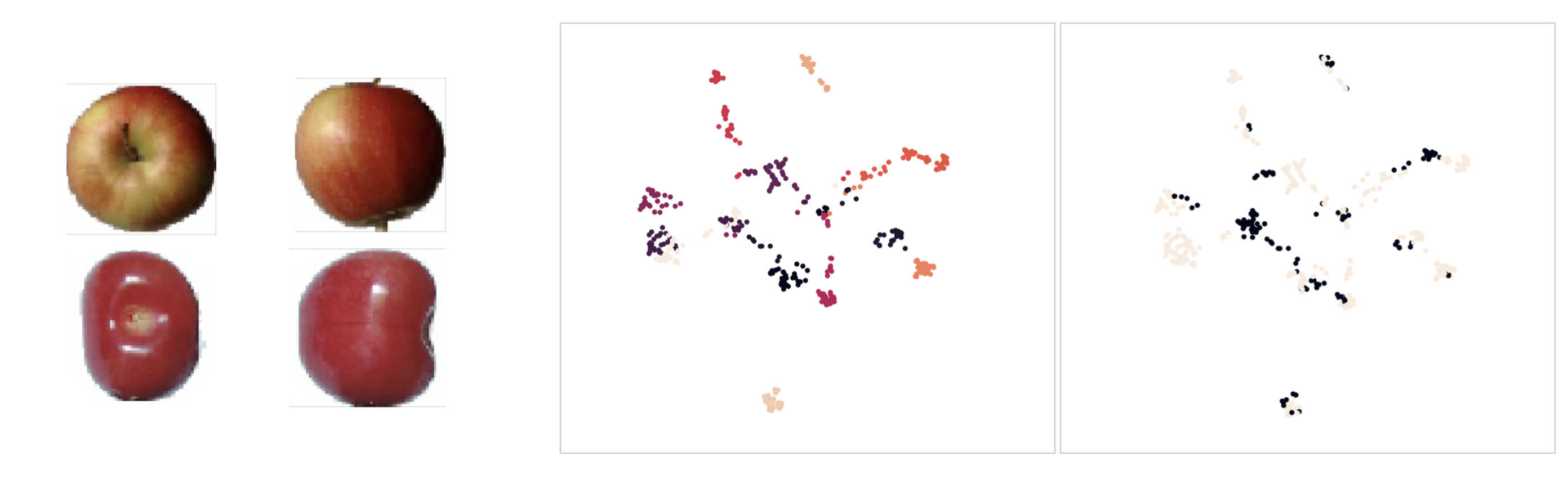} 
\caption{(Left) Example images from Fruits 360. (Center) A low-dimensional visualization of the image of the Fruits 360 dataset in the feature space of a convolutional neural network. Points are colored by type of fruit. (Right) The same visualization with points colored by whether the fruit stem (or stem node) is facing the camera or not.
}
\label{fig-feature-space-visualization}
\end{figure*}

\begin{figure*}[t]
\centering
\includegraphics[width=.9\columnwidth]{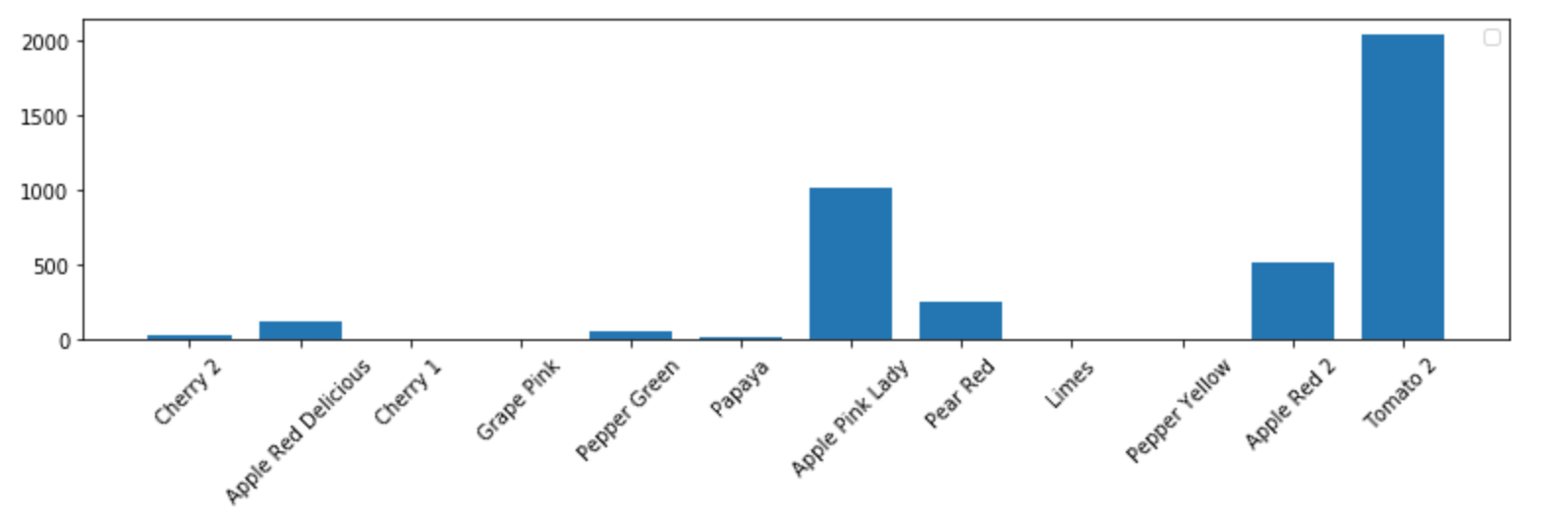} 
\caption{The number of times the removal of each type of fruit causes the maximum increase in ``clusterability'' with respect to \eqref{eqn-protonet-objective-function}.}
\label{fig-sns-plot}
\end{figure*}

\omitt{
\begin{table*}
{\def\arraystretch{1.5}%
\begin{center}
\footnotesize
    \begin{tabular}{ccc}
        $U$ & local inconsistency & $U\setminus V$\\
        \hline
        $\{\text{Pepper Green,Lime}\}$ & 0.367 & $\{\text{Lime}\}$\\
        $\{\text{Papaya, Apple Red Delicious}\}$ & 0.354 & $\{\text{Lime}\}$\\
        $\{\text{Pepper Green, Cherry 1}\}$ & 0.352 & $\{\text{Cherry}\}$\\
        $\{\text{Pepper Green, Pear Red}\}$ & 0.334 & $\{\text{Cherry 1}\}$\\
        $\{\text{Cherry 2, Pepper Green}\}$ & 0.328 & $\{\text{Pear Red}\}$\\
        $\{\text{Pepper Green, Apple Red 2}\}$ & 0.322 & $\{\text{Pepper Green}\}$\\
        $\{\text{Papaya, Grape Pink}\}$ & 0.319 & $\{\text{Apple Red 2}\}$\\
        $\{\text{Cherry 2, Pepper Green, Apple Pink Lady, Apple Red, Cherry 1}\}$ & 0.315 & $\{\text{Grape Pink}\}$\\
        $\{\text{Papaya, Apple Red Delicious, Apple Pink Lady, Pear
Red, Apple Red 2, Cherry 1}\}$ & 0.312 & $\{\text{Cherry 1}\}$\\
        \hline
    \end{tabular}
\end{center}}
\caption{Top $10$ open sets with the highest local $1$-filtered inconsistency}
\label{ex-incoherent-assignment}
\end{table*}}

Many machine learning models built to perform classification tasks can be decomposed into two functions, (1) a {\emph{feature extractor}} that takes a particular data type as input and extracts features most relevant to a task and (2) a simpler {\emph{prediction function}} that uses these features to make predictions. A cat/dog classifier for example might consist of a convolutional neural network feature extractor, $\varphi: X \rightarrow \mathbb{R}^d$, and a predictive function, $\psi: \mathbb{R}^d \rightarrow \mathbb{R}^2$ (where one of the dimensions of $\mathbb{R}^2$ corresponds to the likelihood that the image is a dog and the other corresponds to the likelihood that the image is a cat). $\varphi$ takes an image $x$ from an image space $X$ and encodes it as a vector in $\mathbb{R}^d$, translating the complex visual characteristics of cats and dogs into geometric structure in $\mathbb{R}^d$ that can then be separated by $\psi$. 

When $\varphi$ is a model with relatively few parameters and $X$ is low-dimensional, there exist tools to probe the processes by which the complete model $\psi \circ \varphi$ makes its decisions. However, in many state-of-the-art examples, $\varphi$ is a large (for example over $20$ million parameters), nonlinear model which has been trained on a large, high-dimensional dataset (the benchmark image classification dataset ImageNet \cite{russakovsky2015imagenet} for example contains $1$ million labeled images belonging to $1,000$ different classes of objects). Furthermore the feature space $\mathbb{R}^d$ of large feature extractors tend to themselves be high-dimensional, for example $d = 2048$ for the commonly used ResNet50 convolutional neural network \cite{resnet2016}. Together all of these factors make current deep learning-based machine learning models difficult to analyze. At the same time, because these models are being used in a broader and broader range of applications the need for tools by which to analyze them has become increasingly pressing. In this section we give a very simple example of how the sheaf framework introduced in Sections \ref{sect-data-sheaf}, \ref{sect-model-presheaf}, and \ref{sect-local-consistency} can be used to answer questions about a deep learning-based feature extractor. Our example is deliberately of a different flavor than many of those introduced in the rest of the paper in order to both emphasize of versatility of our methods and point toward an area of future research. 

We take a very simple image dataset, {\emph{Fruits 360}} \cite{murecsan2018fruit}, which consists of images of fruit photographed from various angles (see the left image in Figure \ref{fig-feature-space-visualization} for an example). The standard task associated with this dataset is to predict what type of fruit is in each image, for example, {\emph{Apple}}, {\emph{Cherry}}, etc. Surprisingly, if a feature extractor $\varphi$ is large enough and has been exposed to enough image data it can easily extract the features necessary to differentiate between fruit even if $\varphi$ has never been explicitly trained on this task. One observation made in \cite{kvinge2020fuzzy} is that one can also naturally classify the images in this dataset based on whether or not the stem (or stem node) of the fruit is facing the camera or not. While humans can easily perform this task, the well-trained feature extractor $\varphi$ mentioned above fails here. As described in \cite{kvinge2020fuzzy}, this is an example of a general trend wherein generic computer vision feature extractors tend to be very good at solving problems related to {\emph{class membership}} (i.e. ``does $x$ belong to class $y$'') but less good at problems involving quantity and orientation. The center and right images in Figure \ref{fig-feature-space-visualization} give a $2$-dimensional visualization of the images of the Fruits 360 dataset in $\mathbb{R}^{2048}$ under a feature extractor map $\varphi$ trained on ImageNet. Comparison of the center image, which is colored by fruit type, and the right image, which is colored by whether the fruit's stem is facing the camera or not, suggests that $\varphi$ strongly clusters by type of fruit while only clustering by stem/no-stem at the local level.

To better understand this limitation of $\varphi$ we can ask the following question \\
\vspace{-3mm}

\noindent{\emph{\textbf{Question:} Is $\varphi$ poor at extracting the features required to solve the stem/no-stem problem for all the images in the Fruits 360 dataset or are there instead certain subpopulations on which $\varphi$ particularly struggles dragging down the global performance?}}\\

\vspace{-3mm}
We can use our sheaf framework and filtered inconsistency to begin to answer this question with respect to type of fruit. Let $\Dataset$ be the Fruits 360 dataset. We let $\BaseSpace$ be an index of the images in $\Dataset$ so that the dataset itself can be written as $\{x_i \;|\; i \in I\}$, where $x_i$ is the image indexed by $i$. We let $B = \{\Subbase_{1}, \dots, \Subbase_{k}\}$ be defined such that, for example, $\Subbase_{j}$ might contain all those elements of $\BaseSpace$ that correspond to images containing the class {\emph{Apple}}. The feature extractor we analyze is a ResNet50 convolutional neural network $\varphi: \mathbb{R}^{224 \times 224 \times 3} \rightarrow \mathbb{R}^{2048}$ that maps $224 \times 224$ RGB images to vectors in $\mathbb{R}^{2048}$. We load into $\varphi$ model parameters from the Torchvision library \cite{marcel2010torchvision} that were trained on ImageNet. Thus, $\varphi$ has not been explicitly trained for either the standard Fruits 360 classification task on $\Dataset$ nor the stem/no-stem task. For open set $U \in \Top_B$, let the data sheaf $\DataSheaf$ be defined such that $\DataSheaf(U)$ consists of all functions from $U$ to $\mathbb{R}^{2048}$. Then $\varphi$ defines an assignment of $\DataSheaf$ by defining the function $f_{\Dataset,U}: U \rightarrow \mathbb{R}^{2048}$ by $f_{\Dataset,U}(i) = \varphi(x_i)$.

Our model presheaf will be designed to capture the extent to which subsets of encoded images in $\mathbb{R}^{2048}$ tend to cluster based on their stem/no-stem labels. Inspired by \cite{snell2017prototypical}, we measure the extent to which stem/no-stem images in open set $U$ cluster, in the following way. Suppose $V_1\subset U$ are those elements of $U$ with the ``stem'' label and $V_2 \subset U$ consists of those elements of $U$ with the ``no stem'' label, so that $V_1 \cup V_2 = U$. We randomly draw three examples $i^s_1, i^s_2, i^s_3$ from $V_1$ and three random examples $i^{ns}_1, i^{ns}_2, i^{ns}_3$ from $V_1$. We use these to form {\emph{prototypes}} for those elements of $V_1$ and $V_2$:
\begin{equation*}
\gamma_{s} := \frac{1}{3} \sum_{j=1}^3 f_{\Dataset,U}(i^{s}_j) \quad \text{and} \quad \gamma_{ns} := \frac{1}{3} \sum_{j=1}^3 f_{\Dataset,U}(i^{ns}_j).
\end{equation*}
We predict the stem/no-stem label of the remaining elements $i \in U$ by solving the optimization problem:
\begin{equation} \label{eqn-protonet-objective-function}
\argmin_{r = s,ns} ||\gamma_r - f_{\Dataset,U}(i)||.
\end{equation}
We perform this process several times, recording our accuracy each time. We denote our average accuracy over many trials by $\alpha_{f,U}$. Note that $\alpha_{f,U}$ being closer to $1$ indicates that $f_{\Dataset,U}$ more strongly clusters images based on the stem/no-stem property, since this means that more stem/no-stem labels can be predicted based on their proximity to prototypes for these classes $\gamma_{s}$ and $\gamma_{ns}$.

Our model presheaf $\ModelSheaf$ is designed to store the values $\alpha_{f,U}$ and hence we assign to each $U \in \Top_B$ the closed interval $[0,1]$. Our model map $\ModelMap: \DataSheaf \rightarrow \ModelSheaf$ is defined such that $f_{\Dataset,U} \in \DataSheaf(U)$ is mapped to $\alpha_{f,U}$. Informally, $\ModelMap$ sends the encoding of elements of $U$ in $\mathbb{R}^{2048}$ to a score (based on \eqref{eqn-protonet-objective-function}) that measures how well this encoding captures stem/no-stem clustering.

Because elements of the subbasis $B = \{U_{1},\dots,U_{k}\}$ are mutually disjoint and their union is equal to $\BaseSpace$, $\Top_B$ can be realized as the set of unions of all possible subsets of the sets in $B$, $\Top_B = \{ \cup_{W \in T} W \; | \; T \subseteq B\}$. Let $U \in \Top_B$. Then $U = \Subbase_{j_1} \cup \dots \cup \Subbase_{j_r}$ where $j_1, \dots, j_r \in \{1,\dots,k\}$. Then 
\begin{equation*}
\Lambda^1_U = \{V \in \Top_B \; | \; V = \Subbase_{j_1} \cup \dots \cup \widehat{\Subbase_{j_t}} \cup \dots \cup \Subbase_{j_r}, \;\; 1 \leq t \leq r \}
\end{equation*}
where $\widehat{\cdot}$ denotes omission from the union. Thus for any open set $U$ with $U \neq \emptyset$, $\Lambda^1_U$ consists of all those sets obtained by removing one type of fruit (that is, one $U_{j}$) from the union of subbasis elements that form $U$. 

When calculating the local $1$-filtered inconsistency for each $U \in \Top_B$, we can not only record the $1$-filtered inconsistency itself, but can note the $V \subset U$ at which the max \eqref{eqn-local-inconsis-filtered} is achieved. Then $U\setminus V$ will correspond to a type of fruit. In Figure \ref{fig-sns-plot} we display a bar plot that shows the number of times each type of fruit appears when calculating this statistic. Rather than all types of fruit being equally problematic, we see that there are a few types that most frequently cause the largest drop in accuracy when they are included in the model's evaluation: {\emph{Tomato 2}}, {\emph{Apple Pink Lady}}, {\emph{Apple Red 2}}, and {\emph{Pear Red}}. It is not at all clear why the model tends not to cluster these fruits by their orientation. While {\emph{Tomato 2}} does have a less visually distinctive stem node, {\emph{Apple Pink Lady}} has not only a stem node but a stem itself which one would think a convolutional neural network would capture during feature extraction. 

Note that our method is able to pick up on higher-order effects that result from the interactions of several specific fruit types. For example, perhaps the model is able to cluster stem/no-stem for {\emph{Apple Red 2}} or {\emph{Pear Red}} individually, but because of the way each is represented in $\mathbb{R}^{2048}$ the model struggles when they are together. For this reason, in the future it would be interesting to also examine $\Lambda_U^2$ and $\Lambda_U^3$.

\section{Conclusion and Future Work}

As statistical models grow ever bigger and more opaque, developing methods that give insight into not only the global behavior of the model, but also the local behavior becomes ever more important. In this work we develop a sheaf-theoretic framework to evaluate the fit of data-driven models. We show how a topology can be used to capture various subpopulations of interest within a dataset and then attach statistics related to model fit to each of these subpopulations. Via the notion of inconsistency, we can establish regions of the dataset on which a model's behavior changes significantly. We see application to real datasets as the next critical step in this direction. We expect that the tools developed here will need to be refined not only to meet real-world computational requirements, but also to highlight the aspects of model fit that the model builder actually cares about. Nevertheless, we hope that this work will increase the likelihood that sheaves will be a tool in some future model builder's toolbox.

\bibliography{generic}

\section{Appendix}
\label{appendix}

Proof of Proposition \ref{prop-inconsistency-presheaf}.

\begin{proof}
Suppose that $\ModelMap$ is a presheaf morphism and $U \in \Top_B$ is an open set. Let $F = (f_V)_{V \in \Top_B}$ be a consistent assignment. Then for any $V \subseteq U$, $\res^{\DataSheaf}_{U,V}(f_U) = f_V.$  Since $\ModelMap$ is a presheaf morphism, then by definition for any $f_U \in \DataSheaf(U)$,
\begin{equation*}
\res^{\ModelSheaf}_{U,V}(\ModelMap_U(f_U)) = \ModelMap_V(\res_{U,V}^\DataSheaf (f_U)).
\end{equation*}
But then since $A$ is consistent, $\ModelMap_V(\res_{U,V}^{\DataSheaf}(f_U)) = \ModelMap_V(f_V)$, so $\res^\ModelSheaf_{U,V}(\ModelMap_U(f_U)) = \ModelMap_V(f_V)$ and therefore
\begin{equation*}
d_V(\res^\ModelSheaf_{U,V}(\ModelMap_U(f_U)), \ModelMap_V(f_V)) = 0.
\end{equation*}
Since this is true for all open $V \subseteq U$, then $\Incon(F,U) = 0$.

Now suppose that for any consistent assignment $F$ of $\DataSheaf$, the local inconsistency of $F$ at any open set $U$ is $0$. We need to show that for any $f_U \in \DataSheaf(U)$ and any open $V \subseteq U$, $\res^\ModelSheaf_{U,V}(\ModelMap_U(f_U)) = \ModelMap_V(\res^\DataSheaf_{V,U}(f_U))$. 

Choose some $y \in Y$. Let $\widetilde{f}_I: \BaseSpace \rightarrow Y$ be the function in $\DataSheaf(\BaseSpace)$ that maps
\begin{equation*}
\widetilde{f}_\BaseSpace(i) := \begin{cases}
f_U(i) & \text{if $i \in U$}\\
y & \text{otherwise.}
\end{cases}
\end{equation*}
$\widetilde{f}_\BaseSpace$ defines a consistent assignment $(\res^\DataSheaf_{\BaseSpace,U}(\widetilde{f}_\BaseSpace))_{U \in \Top_B}$. In particular $f_U = \res^\DataSheaf_{\BaseSpace,U}(\widetilde{f}_\BaseSpace)$. Then $F$ has inconsistency $0$ by assumption and hence
\begin{equation*}
\max_{W \in \Lambda_U} d_W(\res^\ModelSheaf_{U,W}(\ModelMap_U(f_U)),\ModelMap_W(f_W)) = 0.
\end{equation*}
In particular, this implies that $d_V(\res^\ModelSheaf_{U,V}(\ModelMap_U(f_U)),\ModelMap_V(f_V)) = 0$ and $f_V = \res^\DataSheaf_{U,V}(f_U)$ hence
\begin{equation*}
\res^\ModelSheaf_{U,V}(\ModelMap_U(f_U)) = \ModelMap_V(\res^\DataSheaf_{V,U}(f_U))
\end{equation*}
Since this is true for all open sets $U,V$ with $V \subseteq U$ and all $f_U \in \DataSheaf(U)$, then $\ModelMap$ is a presheaf morphism.
\end{proof}

\end{document}